\pdfoutput=1
\documentclass[11pt]{article}

\usepackage{acl}

\usepackage{times}
\usepackage{latexsym}
\usepackage{amsmath}

\usepackage[T1]{fontenc}
\usepackage[utf8]{inputenc}
\usepackage{microtype}

\usepackage{subfigure}
\usepackage{inconsolata}

\usepackage{graphicx}
\usepackage{subfigure}

\usepackage{array}
\usepackage{color}

\usepackage{algorithm}
\usepackage{algorithmic}

\usepackage{times}
\usepackage{latexsym}
\usepackage{todonotes}
\usepackage{makecell}
\usepackage{booktabs}
\usepackage{array}
\usepackage[inline]{enumitem}

\usepackage{microtype}

\usepackage{inconsolata}
\title{\texttt{Knowing What LLMs DO NOT Know:} A Simple Yet Effective \\ Self-Detection Method}

\author{
Yukun Zhao\textsuperscript{\rm 1,2} \quad Lingyong Yan\textsuperscript{\rm 2} \quad Weiwei Sun\textsuperscript{\rm 1} \quad Guoliang Xing\textsuperscript{\rm 2} \quad Chong Meng\textsuperscript{\rm 2} 
\\
 \textbf{Shuaiqiang Wang}\textsuperscript{\rm 2} \quad \textbf{Zhicong Cheng}\textsuperscript{\rm 2} \quad \textbf{Zhaochun Ren}\textsuperscript{\rm 3}\thanks{~~Co-corresponding authors.} 
\quad \textbf{Dawei Yin}\textsuperscript{\rm 2}\footnotemark[1] \\ 
\textsuperscript{\rm 1}Shandong University, Qingdao, China \quad
\textsuperscript{\rm 2}Baidu Inc., Beijing, China \\ \textsuperscript{\rm 3}Leiden University, Leiden, The Netherlands\\
\texttt{\{zhaoyukun02,yanlingyong\}@baidu.com,sunnweiwei@gmail.com}\\
\texttt{\{xingguoliang,mengchong01,wangshuaiqiang,chengzhicong01\}@baidu.com} \\
\texttt{z.ren@liacs.leidenuniv.nl,}~\texttt{yindawei@acm.org}
}

\begin{document}
\maketitle
\begin{abstract}

Large Language Models (LLMs) have shown great potential in Natural Language Processing (NLP) tasks.
However, recent literature reveals that LLMs hallucinate intermittently, which impedes their reliability for further utilization. 
In this paper, we propose a novel self-detection method to detect which questions an LLM does not know.
Our proposal is empirical and applicable for continually upgrading LLMs compared with state-of-the-art methods. 
Specifically, we examine the divergence of the LLM's behaviors on different verbalizations for a question and examine the atypicality of the verbalized input. 
We combine the two components to identify whether the model generates a non-factual response to the question. 
The above components can be accomplished by utilizing the LLM itself without referring to any other external resources. 
We conduct comprehensive experiments and demonstrate the effectiveness of our method for recently released LLMs involving Llama 2, Vicuna, ChatGPT, and GPT-4 across factoid question-answering, arithmetic reasoning, and commonsense reasoning tasks.

\end{abstract}

%缺少过度，为什么这么做
%Specifically, we first diversify the textual verbalizations for a given question. Then we examine the divergence between the generated answers corresponding to these verbalized questions. We further compute the likelihood of the input to quantify its atypicality. 

\section{Introduction}
\label{sec:introduction}

\begin{table*}[htbp]
  \centering
  \resizebox{0.99\textwidth}{!}
{
    \begin{tabular}{p{1.35\columnwidth}p{0\columnwidth}p{1.35\columnwidth}}
    \toprule
    \multicolumn{3}{l}{\textbf{Knowledge Quiz}: The lyricist of \emph{Kadam Kadam Badhaye Ja Khushi ke Geet Gaye Ja}} \\
    \hline
    Q1: \texttt{Who has written the lyrics to patriotic song \emph{kadam kadam badhaye ja khushi ke geet gaye ja}?} &       & Q2: \texttt{Who is the lyricist of the patriotic song \emph{Kadam Kadam Badhaye Ja Khushi ke Geet Gaye Ja}?} \\
    R1: \texttt{The lyrics of the patriotic song \emph{Kadam Kadam Badhaye Ja Khushi Ke  Geet Gaye Ja} were written by \textbf{\textcolor{red}{Shri Ravi Shankar Sharma}}.} &       & R2: \texttt{The lyricist of the patriotic song \emph{Kadam Kadam Badhaye Ja Khushi ke Geet Gaye Ja} is \textbf{\textcolor{red}{Shri Pyarelal Santoshi}.}} \\
    \hline
    \multicolumn{3}{l}{Correct Answer:\ \textbf{\textcolor{blue}{\texttt{Vanshidhar Shukla}}} .} \\
    \toprule
    \multicolumn{3}{l}{\textbf{Math Problem} } \\
    \hline
    Q1: \texttt{Tom's restaurant gets 6 reservations a night. They normally order 2 meals that cost \$5 each and a \$5 bottle of wine. How much do they make a week if they are open 2 days a week?} &       & Q2: \texttt{Kanan's restaurant gets 6 reservations a night. They normally order 2 meals that cost \$5 each and a \$5 bottle of wine. How much do they make a week if they are open 2 days a week?} \\
    R1: \texttt{They make 6*2=12 meals a night. So they make 12*10 =120 a night. \textbf{\textcolor{red}{That means they make 120*2=240 a week.}} } &       & R2: \texttt{They make 2*6=12 meals a night. So they make 12*2=24 on meals. They also make 6*5=30 on wine. So they make 24+30=54 a night. \textbf{\textcolor{red}{That means they make 54*2=108 a week.}} }  \\
    \hline
    \multicolumn{3}{l}{Correct Answer:\ \textbf{\textcolor{blue}{\texttt{180}} }.} \\
    \bottomrule
    \end{tabular}%
}
\caption{Two examples of completely different responses for the different verbalized but semantically equivalent questions.}
\label{tab:intro_2_sample}
\vspace{-0.4cm}
\end{table*}

%% 交代背景和问题
With the significant improvements in large language models (LLMs) such as PaLM~\cite{chowdhery2022palm}, ChatGPT~\cite{ouyang2022training}, GPT-4~\cite{OpenAI2023GPT4TR}, LLAMA 2~\cite{touvron2023llama}, and Vicuna~\cite{vicuna2023}, LLMs have been applied in various natural language tasks.
Unfortunately, LLMs still produce unexpected falsehoods~\cite{bang2023multitask, li2023halueval}, i.e., they are unaware of what they do not know and generate responses indiscriminately.
For example, ChatGPT generates falsehoods for a knowledge quiz and math problem, as shown in Table~\ref{tab:intro_2_sample}.
These intermittent errors can severely hinder the LLMs' reliability in practice,
which makes detecting what they do not know an important research problem~\cite{hendrycks2021unsolved,lin2022teaching, kadavath2022language}.

\begin{figure}[t]
    \centering
    \includegraphics[width=1.0\linewidth]{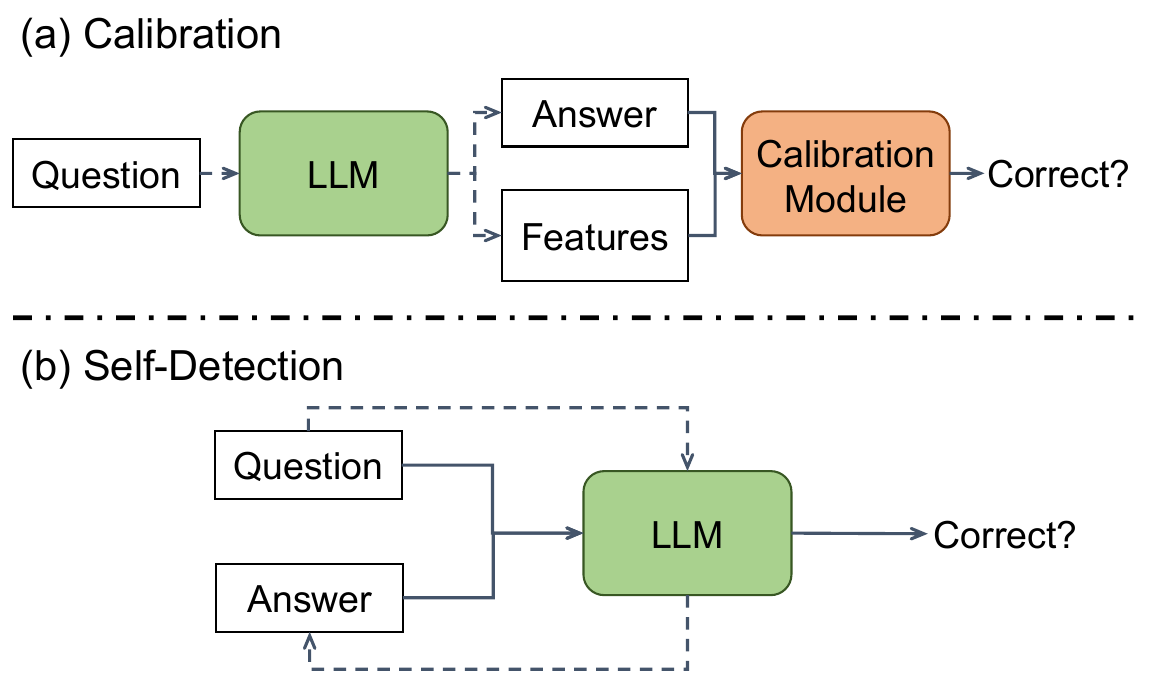}
    \caption{Two paradigms for detecting hallucinations. The dashed lines denote the LLM generation process. The solid lines denote non-factuality detection.}
    \label{fig:paradigm}
    \vspace{-0.5cm}
\end{figure}

There are two main paradigms to detect non-factuality: the calibration-based and the self-detection methods.
The first class of methods calibrates the model confidence to better detect falsehoods of the generations (See Figure~\ref{fig:paradigm}(a)).
Among them, \citet{mielke2022reducing} train auxiliary calibrators, \citet{lin2022teaching} and \citet{jiang2021can} improve the calibration through fine-tuning the language model. We propose a self-detection method that does not require further fine-tuning. 
% or referring to any other external resources.
% Nevertheless, these methods are mostly ad-hoc, while the score may not be available for commercial LLMs, and the tuned calibration is challenging for the generalization of out-of-distribution data.
%The confidence scores are generating probabilities~\cite{jiang2021can,si2022prompting,manakul2023selfcheckgpt} or calibrated probabilities using auxiliary models~\cite{lin2022teaching, jiang2021can}.

The self-detection methods directly leverage the LLMs themselves to detect whether they hallucinate (See Figure~\ref{fig:paradigm}(b)).
For example, ~\citet{kadavath2022language} prompt the LLMs to predict the confidence score on whether their responses are true, and ~\citet{si2022prompting} directly utilizes the token probabilities of the generations as the confidence score; \citet{wang2022self} and \citet{manakul2023selfcheckgpt} detect the falsehoods by sampling answers with a high temperature and examining self-consistency between them. 
However, the performance of these works is limited as LLMs tend to be overconfident about their own outputs and these work would be less effective after the models are trained more aligned~\cite{OpenAI2023GPT4TR,ouyang2022training,zhao2022calibrating}.

%the model would provide a random response, sometimes correct while other times incorrect for a question

% The reasons that LLMs produce incorrect responses can be classified into two types. The first one is that the trained model has stored falsehoods, and we need to refer to external knowledge to identify these errors~\cite {kryscinski2019evaluating, wang2020asking, yin2023large}. The other one is the model has a weak grasp of the knowledge or has no knowledge at all.
% Ideally, if an LLM knows a certain knowledge, it should be able to provide correct answers regardless of the way the questions are verbalized, i.e., not limited to specific question formats. When a model provides completely different responses for the same questions with different verbalization, as shown in Table~\ref{tab:intro_2_sample},  we consider that the model does not know the question, and our paper aims to identify this unknown situation.
%That means if the LLMs have hallucinations about a given question, they tend to provide different responses to the questions in different verbalizations but with the same semantics.
%Therefore, if the LLMs do not know the corresponding answers, they usually tend to give different answers if we ask them the questions in different expressions but with the same semantics.

% based on biased context.
In this paper, we consider detecting non-factuality as that a model does not know which knowledge is related to the question or does not understand the queried question, outputting the non-factual response. 
A model is expected to provide correct and consistent answers regardless of the ways the questions are verbalized. Therefore, if it responds drastically differently to the different verbalizations, we consider the model does not know the question.

Built on the above hypothesis, we propose a novel self-detection method that includes 1) examining the divergence of the LLM's behaviors on different verbalized questions and 2) examining whether the verbalization of the question is typical in the LLM as shown in Figure~\ref{fig:framework}.
% this paper proposes a novel self-detection method for LLMs' non-factuality, which examines the divergence of LLMs' behaviors on different verbalizations of the question. Moreover, we examine the verbalized 
%given a question about a piece of undecided knowledge or facts,
Specifically, for the first component, we first diversify the queried question to several semantically equivalent verbalizations.
Then, we examine the divergence between the answers corresponding to the questions.
For the second component, we use the negative log-likelihood of the verbalized question as the indicator of atypicality in the language model. 
Concurrent work~\cite{zhang2023sac} has also mentioned rephrasing the original question to alternatives and checking the consistency of the answers with the original answer. In contrast, we further propose to examine the representativeness of the input for the model and examine the divergence in the answer distribution. Our self-detection method is applicable for continually upgrading LLMs.

To verify the effectiveness of our method, we conducted extensive experiments on GPT-4, ChatGPT, Vicuna, and Llama 2 across three types of tasks: factoid question answering, commonsense reasoning, and arithmetic reasoning tasks.
The experimental results demonstrate the superior performance of our self-detection method.

In summary, our contributions are as follows:
\begin{itemize}
\item We show existing LLMs intermittently retain the verbalization-sensitive problem, generating drastically contradicting responses to the questions with the same semantics but verbalized differently.
                               
%\item We utilize the divergence of the LLM's responses across various verbalizations to detect which parts of knowledge LLMs are unknown.

\item We introduce a self-detection suit that relies solely on an LLM itself, enabling a light detection of whether an LLM is unknown for a question.

% \item Our experiments reveal that the methods based on generated probabilities perform the worst, and the performance of self-consistency decreases when the models are more aligned.

\item We prob what an LLM knows and does not know and show a correlation between the unknown to the popularity, the reasoning steps, and the formulations.

\end{itemize}

\section{Related Work}
\label{sec:related_work}

\paragraph{Model Calibration} 
% 按照方法组织 + 校准的模型
Calibration is a well-studied topic in traditional neural networks~\cite{hendrycks2016baseline, guo2017calibration, pereyra2017regularizing, qin2021improving}, aiming to provide a confidence score that aligns well with the true correctness likelihood. 
~\citet{jagannatha2020calibrating}, \citet{jiang2021can} and \citet{kadavath2022language} show BERT~\cite{devlin2018bert}, DistilBERT~\cite{sanh2019distilbert}, T5~\cite{raffel2020exploring}, BART~\cite{lewis2019bart}, GPT-2~\cite{radford2019language}, GPT-3.5~\cite{ouyang2022training} are not well-calibrated on the language tasks.

Post-hoc methods like temperature scaling and feature-based fitting on a development set are widely used~\cite{guo2017calibration, desai2020calibration, hendrycks2019using, jiang2021can}, which are straightforward to implement.
Bootstrapping and ensembling methods~\cite{osband2016deep, lakshminarayanan2017simple, sun2022quantifying,radford2019language} are explored for the traditional DNN models.
\citet{li2022calibration,ye2022can,dong2022calibrating,yuksekgonul2023beyond} fine-tune and optimize the calibration for BERT, RoBERTa, T5 and Alpaca respectively.
%\citet{li2022calibration} utilize token attribution to the classification to optimize the calibration loss on BERT~\cite{devlin2018bert} and RoBERTa~\cite{liu2019roberta} for the NLI, paraphrase detection, common sense reasoning tasks. \cite{ye2022can} train a calibrator with explainable features for RoBERTa~\cite{liu2019roberta}. ~\citet{dong2022calibrating} calibrate factual knowledge generation in T5~\cite{raffel2020exploring} via contrastive knowledge assessment. 
\citet{mielke2022reducing} and \citet{lin2022teaching} fine-tune the BlenderBot~\cite{roller2020recipes} and GPT-3~\cite{brown2020language} separately for calibration and express the models' uncertainty in a verbalized statement. 
The calibration tuned for specific tasks makes it challenging to generalize on out-of-distribution data~\cite{guo2017calibration}.

\paragraph{Hallucination Detection} 
% LLM有哪些能力，生成的好（more consistency）、self-instruct、self-consistency、self-know
LLMs such as ChatGPT~\cite{ouyang2022training}, GPT-4~\cite{OpenAI2023GPT4TR}, Vicuna~\cite{vicuna2023}, Llama 2~\cite{touvron2023llama} and Claude~\cite{Claude} have obtained remarkable performance on various language tasks~\cite{bang2023multitask, rangapur2023chatgpt}. 
%such as they generate high-quality human-like text in response to natural language inputs~\cite{rao2023can}, and have the great potential for translation~\cite{jiao2023chatgpt}, NLI, paraphrasing~\cite{zhong2023can}, summarization~\cite{wang2023cross}, coding~\cite{zhuo2023large}, optimization, layout issues~\cite{michail2023uzh_clyp}, conversational QA tasks~\cite{rangapur2023chatgpt} and reasoning tasks~\cite{wang2022self} with proper prompts~\cite{brown2020language, kojima2022large, wei2022chain}. 
However, %recent work~\cite{madaan2023self, fu2023improving} show that GPT-3.5~\citep{ouyang2022training,welleck2022generating} and GPT-4~\cite{OpenAI2023GPT4TR} can understand the feedback on their own generated response instructed from evaluators or humans and improve their output interactively in tasks like social media conversations, code optimization, and constrained generation. 
recent work~\cite{mallen2022not, bang2023multitask, li2023halueval,yin2023large} show that LLMs may produce hallucinated contents, i.e., non-factual responses. The importance of the hallucination problem has been highlighted by several work~\cite{lin2022teaching,ji2023survey} as it hinders the reliability of the LLMs.

%\paragraph{Uncertainty and Hallucination}
%1. 比对references or external knowledge
%2 token probability
%3 muliti-answers, 包括rephrase answers和uncertainty
%4  
\citet{kadavath2022language} and \citet{agrawal2023language} use LLMs to evaluate the sampled answers but can not evaluate their self-generated answers due to overconfidence. \citet{si2022prompting} and \citet{manakul2023selfcheckgpt} utilize their confidence scores like token probability to indicate the confidence of their output.
Recent work~\cite{wang2022self, si2022prompting,mundler2023self, kuhn2023semantic} examines the self-consistency score among the randomly sampled answers which are generated through a higher temperature. Both the confidence score of the model output and sample-based score highly rely on the current model training, which means the methods would not be that effective after the models are trained to be more aligned. 

~\citet{llmscan} combine the LLMs verbalized statement, self-consistency of the randomly sampled answers, and the consistency between the induced answers. This work proposes to add additional instruction to the prompt for generating induced answers. 
Concurrent work \cite{zhang2023sac, cohen2023lm} utilizes several verifier LLMs to cross-check whether a language model generates falsehoods.
~\citet{zhang2023sac} also rephrases the original question to alternative inputs and checks the consistency of the answers with the original answer as the confidence score. We propose a unified method that examines the divergence of the LLMs' behaviors across the diversified questions besides the consistency pair and the atypicality of the verbalized input in the LLMs. Our proposal is self-detection without referring to any other LLMs or external resources.

%\citet{amayuelas2023knowledge} propose a known unknowns dataset and show a minimal gap in the answer uncertainty between known and unknown questions in their dataset. \citet{agrawal2023language} ask additional questions about the generated output and check the consistency of additional answers to detect the hallucinated contents. 
 
%If we increase randomness, we certainly obtain different answers. When using the contradiction between these answers to detect LLMs' unknown, it would be confusing whether the randomness or the LLMs' actual unknown contributes to the results.

\section{Inconsistency and Atypicality in LLMs}
\label{sec:background}

We attribute the non-factuality of an LLM to the generative characteristics which sample the most possible tokens sequentially.
It means even if the LLM does not know the exact knowledge related to the question or even does not understand the question (understand the instruction), it still generates plausible responses as observed in previous work~\cite{cao-etal-2021-knowledgeable, zhuo_etal_2023_red_teaming_chatgpt_via_jailbreaking:_bias_robustness_reliability_and_toxicity}.

Consequently, if an LLM returns contradicting responses to the semantically equivalent questions, the LLM does not know the knowledge for the question. Besides, if the textual verbalization of a question is not representative in the LLM, i.e., atypical, it would be hard to understand resulting in a lower-quality response~\cite{yuksekgonul2023beyond}.
Two examples of ChatGPT are shown in Table~\ref{tab:intro_2_sample}, where the {Q1} and {Q2} describe the same question with different verbalizations, but their answers are completely different.

% Based on the observation, this paper proposes to leverage the inconsistency between the responses to different verbalizations of the same questions, to automatically detect the LLMs' non-factuality.
So, we 1) examine the divergence between the responses ($R = \{r_1, ..., r_n
\}$) to a question set ($Q=\{q_1, ..., q_n\}$), where any two questions $q_i$ and $q_j$ are semantically equivalent; 2) then examine whether the verbalized question $q$ is representative in the LLM using the atypicality $A(q)$ of the input.

% \paragraph{Inconsistency Definition} We then check whether the responses ($R = \{r_1, ..., r_n
% \}$) to a question group ($Q=\{q_1, ..., q_n\}$) are inconsistent, where any two questions $q_i$ and $q_j$ are semantically equivalent.
% The inconsistency score is inferred from the group $R$.

% \paragraph{Atypicality Definition} We examine the atypicality of the input $A(q)$ to indicate whether the question $q$ is underrepresented for the LLM.

\section{Self-Detecting What LLMs Do Not Know}
\label{sec:framework}
\begin{figure*}[!ht]
    \centering
    \includegraphics[width=0.98\linewidth]{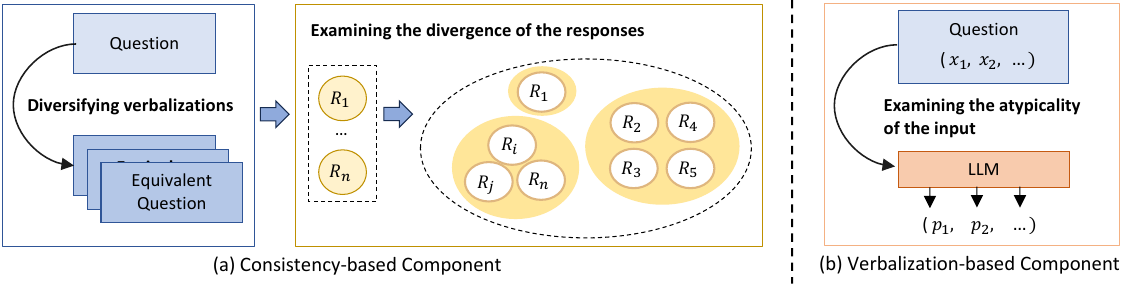}
    \vspace{-0.1cm}
    \caption{The framework of self-detecting what language models do not know.}
    \vspace{-0.1cm}
    \label{fig:framework}
    \vspace{-0.0cm}
\end{figure*}

In this section, we introduce our framework including consistency-based detection~\ref{subsec:consistence-detetion} and verbalization-based detection~\ref{subsec:verbalized_detec} as shown in Figure~\ref{fig:framework}.  

\subsection{Consistence-based Detection}
\label{subsec:consistence-detetion}
Given a question, we first diversify the original question to several questions (Section~\ref{sec:parapharse_question}).
Then, we examine the consistency among the generated responses corresponding to the diversified questions (Section~\ref{subsec:detecting}).

\subsubsection{Diversifying Question Verbalizations}
\label{sec:parapharse_question}
We diversify question $q$ to several textual verbalizations $Q(q) = \{q_1, ..., q_n\}$ that express the same meaning.

\paragraph{Model-based Generation} For those open QA questions, we exploit a LLMs itself (eg., ChatGPT, Vicuna) to generate paraphrased questions through the prompt: \texttt{Given the following question [QUESTION], paraphrase it to have different words and expressions but is semantically equivalent.}
% Specifically, we set a higher temperature to obtain $m$ paraphrased questions.
The unbroken instruction for the task is shown in Table~\ref{tab:paraphrase_prompt} in Appendix~\ref{sec:prompts}.

After obtaining the paraphrased questions, we filter out the unsatisfied ones by prompting the language model to detect whether two questions are semantically equivalent and the instruction is shown in Table~\ref{tab:filter_prompt}.

\paragraph{Rule-based Generation} For commonsense reasoning and arithmetic reasoning questions, we use expert-defined rules for diversification, as those questions are sensitive to numerical numbers, modifiers, and logical relationships.
We exchange the order of choices provided for the question to obtain $n$ paraphrased questions for commonsense reasoning.
We substitute the person names of a question with new names to obtain $n$ paraphrased questions for arithmetic reasoning problems, as the second example in Table~\ref{tab:intro_2_sample}.

% Finally, we obtain at most $n$ verbalized questions $Q(q)$.

% \subsubsection{Generating Candidate Answers}
% \label{sec:generate_answers}
% After obtaining the diverse verbalized questions, we prompt the LLM to generate its candidate responses according to the questions.
% We employ a greedy decoding strategy to avoid unpredictable randomness of the LLMs as much as possible.

% In this stage, we generate a unique response $r_i$ separately for each question $q_i$ and obtaining the response set $R(q)=\{..., r_i, ..., r_n\}$ for $Q(q)=\{..., q_i, ..., q_n\}$.
% After that, we examine the divergence score between the candidate answers to determine whether the LLM may generate non-factual answers for this question.

\subsubsection{Calculating Consistency Score}
\label{subsec:detecting}
We examine the consistency among the generated responses $R(q)=\{r_1, ..., r_n\}$ according to the diversified questions $Q(q)=\{q_1, ..., q_n\}$. For generation, we employ the LLM using the greedy decoding strategy to avoid unexpected randomness of the generative model as much as possible.
%To obtain the consistency score, we first determine whether any two answers are consistent.

\paragraph{Consistency Determination}
Firstly, we examine whether any two answers are consistent $I(r_i, r_j) \in \{0, 1\}$.
For these answers with fixed formats like multiple-choice answers, we extract the final answer using regular expressions and check whether the final answer exactly matches (EM) the other one.
For these free-form answers, we use the LLM itself to handle the inconsistency detection by asking whether the two answers are the same or contradicting, as shown below. The $I(r_i, r_j)$ is inferred from the generated contents using keywords "Contradicted" or "Same".

\begin{table}[!th]
\small
\vspace{-0.2cm}
\begin{tabular}{p{7.2cm}}
\toprule
\texttt{Determine whether the answer 'A1' is 'Contradicted' or 'Same' with the answer 'A2' for the question 'Q'. You need to check whether the two answers exactly describe the same thing such as the same entity, digit, or arithmetical results. If the two answers are the same, give "Same", otherwise give "Contradicted" as the result.} \\
\bottomrule
\end{tabular}
\vspace{-0.2cm}
\caption{The instruction for determining whether two answers are consistent.}
\label{tab:consistency_prompt}
\vspace{-0.4cm}
\end{table}

This task is a strength of the latest LLMs even in a zero-shot measure as it demands basic logical reasoning abilities~\cite{qin2023chatgpt, liu2023evaluating, zhong2023can} and we conduct the human evaluation for this component at the experiments.

\paragraph{Consistency Calculation}
A common way of calculating the consistency score is:
\begin{equation}
    Consistency(R(q)) = \frac{1}{n-1} \sum_{r_i, r_i\neq r} I(r_i, r)
\end{equation}
where $r$ is the response for the original question $q$.

We further compute the divergency of the response distribution to characterize the uncertainty about the question.
Based on consistency, we group the responses into several clusters and obtain a cluster distribution $\Omega=\{\omega_1, ..., \omega_k\}$ for the $n$ responses. 
Specifically, we perform the following clustering algorithm \ref{alg1}:% ~\ref{alg1}:
\begin{algorithm}
    \caption{Clustering Answers}
    \label{alg1}
    \begin{algorithmic}[1]
    \STATE Input: $R(q), \{I(r_i, r_j)\} $
    \STATE Output: $\Omega = \{\omega_1, ..., \omega_k \}$
    \STATE Initialization: $\omega_1 = \{r_o\} $, where $r_o$ is randomly sampled from $R(q)$
    \FORALL{$r_j \in R(q), r_j \neq r_o$ } 
    % \IF{ $r_i$ == $r_j$ }  
    % \STATE { continue}
    % \ENDIF
    \STATE $Clustered = False $
    \FORALL{ $\omega_l \in \Omega$  } 
    \STATE Randomly draw a response $r_i$ from $\omega_l$
    \IF{ $I(r_j, r_i) == 1$ }  
    \STATE $\omega_l \gets \omega_l+ r_i,  Clustered=True$
    \STATE Break
    \ENDIF
    \ENDFOR
    \IF{ $Clustered == False $ }
    \STATE $\omega_{new}=\{r_j\}, \Omega \gets \Omega + \omega_{new} $ 
    \ENDIF
    \ENDFOR
    \end{algorithmic} 
\end{algorithm}

% \paragraph{Entropy-Based Detection}
After clustering, we calculate the entropy of the answer distribution as another consistency score:
\begin{equation}
    Entropy(R(q)) = \sum_{l} \frac{N(\omega_l)}{n} \log{ \frac{N(\omega_l)}{n}}
\end{equation}
where $N(\omega_l)$ is the number of responses in the cluster $\omega_l$.
The entropy measures the degree of divergence between the responses to the same question. A higher entropy indicates greater randomness in the generations. It corresponds to a lower probability of providing correct answers for the question, which suggests the LLM is less likely to know the question.

\subsection{Verbalization-based Detection}
\label{subsec:verbalized_detec}
We then compute the atypicality of the input. 
Inspired by~\cite{yuksekgonul2023beyond}, current LLMs are autoregressive models that compute a marginal distribution $P (x)$ as its confidence score. We compute the negative log-likelihood of the verbalized input as the indicator of the atypicality:
\begin{equation}
    A(q) = - \log P (q) = - \sum_t^{T} \log P(x_t | X_{<t})
\end{equation}
where $x_t$ and $X_{<t}$ indicate a token and a token set in the question $q$. We add a normalized score $A(q) / N(q)$ in this component, where $N(q)$ is the number of tokens in question $q$. We use $A(q)$ along with its normalized version as the atypicality of the input to quantify whether the verbalized input is representative in the language model. A higher value of $A(q)$ would indicate that the verbalization is more atypical for the language model.

Finally, we combine the two components (detailed in Section~\ref{subsec:setup}) to predict the final confidence score that the LLM does not know the question.
% \subsection{Final Confidence Score}

\section{Experiments}
\label{sec:experiments}

\begin{table*}[!t]
\centering\small
\begin{tabular}{lp{3.5em}<{\centering}p{6.5em}<{\centering}p{6em}<{\centering}p{3.5em}<{\centering}p{3.5em}<{\centering}p{3.5em}<{\centering}p{3.5em}<{\centering}}
\toprule
& ARC & CommonsenseQA & GSM-8K &	SVAMP & FaVIQ & ComQA  \\
\midrule
\multicolumn{4}{@{}l}{\emph{ChatGPT}}\\
Random & 10.78 & 22.49 & 11.77 & 17.94 & 45.96 & 27.05 \\
ConsistAnswers &	14.24 & 25.96 & 52.71 & \textbf{30.50} & 57.09 & 31.76\\
SelfCheckGPT & 23.60 & 39.38 & 21.14 & 25.68 & 52.26 & 39.56 \\
SelfDetection (w/o Atypicality) & \textbf{40.86} & \textbf{40.23} & \textbf{56.29} & 28.18 & \textbf{59.65} & \textbf{42.86} \\
\midrule
% \multicolumn{4}{@{}l}{\emph{GPT-3.5}} \\
% Random & 17.80 &	26.24	& 67.03 &	30.01	& 53.67 & 52.97 \\
% TokenProbs & 17.32 & 32.43 & 68.97 & 43.55	& 58.21 & 68.38 \\
% Perplexity & 17.62 & 32.35 & 71.63 & 44.74 & 59.85 & 70.07 \\
% ConsistAnswers & 31.97 &	42.57 & 70.97 & 43.35 & 64.22 & 76.89 \\
% SelfCheckGPT & 33.92 & 44.97 & 65.86 & 29.51 & 49.17 & 60.53 \\
% SelfDetection & \textbf{51.15} & \textbf{65.83} & \textbf{73.25} & \textbf{46.78} & \textbf{67.63} & \textbf{79.44} \\
% \midrule
\multicolumn{4}{@{}l}{\emph{GPT-4}}\\
Random          & 6.29 & 9.71  & 6.91 &   7.13 & 37.67 & 23.02 \\
ConsistAnswers & 27.44 & 35.47 & 22.39 &  \textbf{25.99} & 51.30 & 37.34 \\
SelfCheckGPT & 21.15   & 39.26 & 12.99 & 22.87 & 46.66 & 46.31 \\
SelfDetection (w/o Atypicality) & \textbf{36.45}  & \textbf{42.71} & \textbf{36.83} & 24.78 & \textbf{56.26} & \textbf{58.95} \\
\midrule
% atypicalities = [48.76, 55.37, 42.83, 60.73, 31.95, 50.29]
% consistency   = [48.23, 59.76, 43.24, 67.85, 30.45, 60.93]
\multicolumn{4}{@{}l}{\emph{Vicuna-13B}}\\
Random          & 35.45 & 51.15	& 35.94   &	54.92   & 31.56 & 35.32 \\
TokenProbs & 40.66 & 52.39 & 39.03   & 60.00  & 34.39 & 59.18 \\
Perplexity & 41.27 & 52.01 & 37.63   & 61.60 & 36.43 & 59.58 \\
ConsistAnswers  & 42.69 & 54.13 & 43.97 &  63.28  & 24.44 & 50.84 \\
SelfCheckGPT    & 40.43 & 54.52 & 36.49 & 60.35   & 18.81 & 26.52 \\
SelfDetection   & \textbf{54.55} & \textbf{62.93} & \textbf{53.31} &   \textbf{71.19}   & \textbf{39.45} & \textbf{66.97} \\
SelfDetection (w/o Atypicality)    & 48.23 & 59.76 & 43.24 & 67.85 & 30.45 & 60.93 \\
SelfDetection (w/o Consistency)    & 48.76 & 55.37 & 42.83 & 60.73 & 31.95 & 50.29 \\
\midrule

\multicolumn{4}{@{}l}{\emph{Llama2-13B}}\\
Random          & 64.27 & 58.93	& 34.25 & 57.43	& 31.44 & 37.27 \\
TokenProbs      & 64.10 & 62.92 & 35.12 & 55.73 & 33.21 & 43.84 \\
Perplexity      & 64.08 & 62.88 & 35.18 & 55.87 & 33.53 & 44.70 \\
ConsistAnswers  & 71.17 & 61.79 & 47.43 & 63.84 & \textbf{59.16} & \textbf{65.34} \\
SelfCheckGPT    & 69.59 & 60.95 & 33.77 & 59.79 & 40.69 & 41.23 \\
SelfDetection  & \textbf{77.73} & \textbf{71.95} & \textbf{50.38} &   \textbf{70.33}   & 39.83 & 52.36 \\
SelfDetection (w/o Atypicality)    & 65.88 & 65.13 & 40.80 & 61.34 & 41.42 & 52.42 \\
SelfDetection (w/o Consistency)    & 70.90 & 64.00 & 38.19 & 62.08 & 34.19 & 40.26 \\
\bottomrule

\end{tabular}
\caption{The PR-AUC of different methods for ChatGPT (gpt3.5-turbo), GPT-4, Vicuna-13B and Llama2-13B on 6 representative datasets of commonsense reasoning, arithmetic reasoning, and question answering tasks. The best results are shown in bold.}
\label{table:overall_performance}
\vspace{-0.5cm}
\end{table*}

\subsection{Experimental Settings}
\label{subsec:setup}
\paragraph{Datasets}
We evaluate the effectiveness of our self-detection on factoid question answering, arithmetic reasoning, and commonsense reasoning tasks. For factoid question answering, we use FaVIQ~\cite{park2021faviq} and ComQA~\cite{abujabal2018comqa} as our benchmark dataset.
For arithmetic reasoning, we use GSM-8K~\cite{cobbe2021training} and SVAMP~\cite{patel2021nlp}.
For commonsense reasoning, we use ARC-Challenge~\cite{clark2018think} and CommonsenseQA~\cite{talmor2018commonsenseqa}.
For FaVIQ, we randomly split the a-set into train, dev and test sets, and samples 500, 500, and 200 instances respectively. For other datasets, we use the built-in splits and sample the same number of instances for training, validating and testing.

% For ARC-Challenge, comQA, SVAMP and CommonsenseQA, we report the performance on the test set.
% For GSM-8K, we report the performance of test set. For FaVIQ, we randomly split the a-set into train, dev and test set, which contains 500, 500, and 200 samples, respectively. We report the performance on the test set.
\paragraph{Models}
We self-detect the SOTA LLMs including ChatGPT (gpt-3.5-turbo), GPT-4, Vicuna-13B and Llama2-13B (Llama2-13B-chat). For GPT-series models, we request the openAI APIs\footnote{\url{https://platform.openai.com/docs/api-reference}} to obtain the responses. We deploy the models Vicuna and Llama 2 ourselves each using two A100 40G GPUs.

\paragraph{Evaluation Metrics} We report PR AUC 
to measure whether our predicting score correlates well with a nonfactual response. 
For each question in the datasets, we have a golden answer. For factoid question answering tasks, we prompt GPT-4 to verify the correctness of the response by comparing it with the golden answer similar to what we described before. For arithmetic and commonsense reasoning questions, we check whether the final answer exactly matches the golden answer, while the final answer is extracted using regular expressions. If the extraction fails, we prompt GPT-4 to assess whether the answer is correct as we did in the factoid question answering tasks. %to extract the answer  By comparing the uncertainty score our method outputs with the unknown label, we report the PR-AUC metric.

\paragraph{Baselines} We compare our self-detection with recent SOTA methods including: 1). Token-level probability (TokenProbs for short), proposed in~\cite{manakul2023selfcheckgpt}, measures the response's likelihood and the average of the token probabilities is used as the confidence score; 2). Perplexity, the reciprocal of the (normalized) language model probability, is used to indicate the uncertainty~\cite{si2022prompting}; 3). Self-consistency of answers (ConsistAnswers for short) is calculated as the consistency of the sampled answers while the answers are sampled using a high-temperature value (0.7) leading to 10 different predictions~\cite{si2022prompting}; 4). SelfCheckGPT~\cite{manakul2023selfcheckgpt} combines the averages of the main response's BERTScore with the most similar sentence of each drawn sample and the token-level probability.

\paragraph{Implementation Details} For paraphrasing, we set a high temperature 1.0 to obtain 10 re-phrasings for each question. We incorporate the 10 re-phrasings for each question and expand the original training sets and validation sets to 10 times larger. To generate the corresponding answers, we use the default template of each model and employ greedy decoding setting temperature 0.0 to avoid unexpected randomness. This decoding strategy still fits for filtering wrong paraphrases and determining consistency. We employ an XGBoost to fit the four features ($Consistency(R(q))$, $Entropy(R(q))$, $A(q)$ and its normalized version $A(q)/N(q)$) in the expanded training sets and choose hyperparameters from the expanded dev sets. The implementation codes are accessible at this URL\footnote{\url{https://github.com/yukunZhao/Self-DETECTION}}. We report the performance of the six original test sets.

\subsection{Overall Performance}
%We first validate the effectiveness of our proposed self-detection method. 
In Table~\ref{table:overall_performance}, we report the overall performance of six methods on ChatGPT, GPT-4, Vicuna-13B, and Llama2-13B across six datasets. Since we cannot obtain the token probabilities for ChatGPT and GPT4, we omit perplexity and token probability methods and only report the performance of SelfDetection without atypicality. 
The random method assigns a score between 0 and 1 randomly denoting whether the generation is nonfactual serving as the lowest baseline for comparison. The PR-AUC values across different models are not comparable. This is because the ground-truth labels of the four models, whether the models know the answer to a question, are not the same as we report the unknown ratios of each model in Appendix~\ref{subsec:evoluation}. We compare different methods within the same model.

We see that compared with recent methods, our self-detection method mostly achieves the best performance on the six data sets, validating the effectiveness of our method on different LLMs.
Specifically, self-detection shows significant improvements for the commonsense reasoning task on ARC and CommonsenseQA, compared to the previous baselines.
In math problems, GSM8k and SVAMP, the self-detection method demonstrates mostly optimal performance, and the consistAnswers serve as a strong baseline. %For arithmetic reasoning, we guess the generated probabilities, like \emph{2 + 3 = ?}, have a strong correlation with the likelihood of the answer being correct, contributing to the performance.
For the two QA datasets FAVIQ and ComQA, the self-detection method performs the best except on Llama 2, and the consistAnswers method serves as a strong baseline.

% Confidence-based methods like TokenProbs and Perplexity perform relatively lower than other methods. Methods based on self-consistency like ConsistAnswers and its varied version SelfCheckGPT performs better and  

Overall, our self-detection achieves the best performance because we capture the essence of identifying what a language model knows. If a question is atypical or the answers for a question are unstable, the probability of its response being coincidentally correct aligns with the consistency level of its responses and its atypicality.

\subsection{Ablation Study}
%We study which parts contribute to detection performance. 
We see the performance of SelfDetection drops when we remove atypicality or consistency indicating the effectiveness of each component in Table~\ref{table:overall_performance}.
We also see the performance drops greater when we remove consistency compared with atypicality in most datasets, which reveals that the divergence between the answers for diversified questions is more crucial for the SelfDetection method. 
Additionally, we see the performance falls behind self-detection in most cases when we combine ConsistAns with atypicality in Table~\ref{table:consistency_performance} in Appendix~\ref{subsec:additional_exp}, which again verifies the benefits of using verbalizations. 

We conduct a simple linear combination of the two components and see the performance is comparable with the XGBoost fitted one in Table~\ref{table:other_combinatioin} in Appendix~\ref{subsec:additional_exp}. The way of combinations of the components is not vital in our method.

\begin{figure}[!ht]
    \centering
    \subfigure[Comparison on Vicuna-13B]{
    \includegraphics[width=0.98\linewidth]{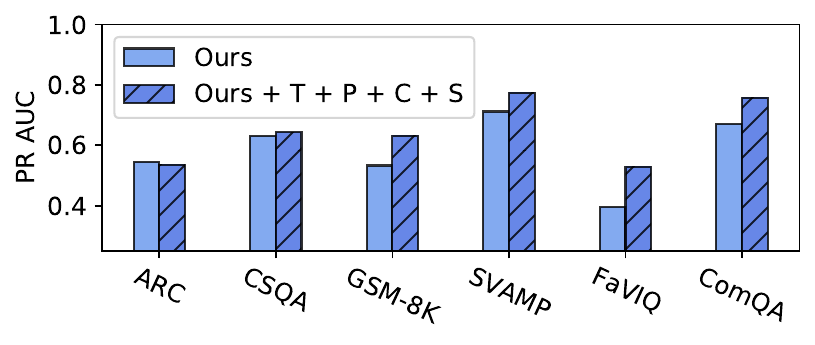}
    \vspace{-0.6cm}
    }
    \subfigure[Comparison on Llama2-13B]
    {\vspace{-0.8cm}
    \includegraphics[width=0.98\linewidth]{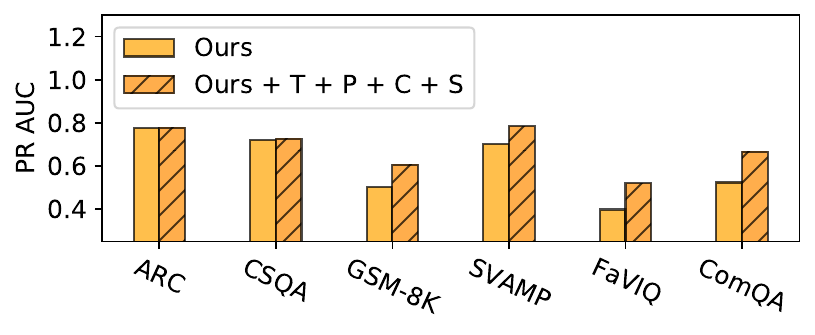}
     \vspace{-0.6cm}}
    \vspace{-0.1cm}
    \caption{The PR AUC when combining our method and previous proposed TokenProbs (T), Perplexity (P), ConsistAnswers (C), and SelfCheckGPT (S). }
    \vspace{-0.0cm}
    \label{fig:ablation}
    % \vspace{-0.0cm}
\end{figure}

Besides, we see the performance is continuously improved when combining our components with the previously proposed tokenProbs, perplexity, consistAnswers, and SelfCheckGPT as shown in Figure~\ref{fig:ablation}. 
We do not report the performance of all combinations of these components as this is not the focus of this paper.

\subsection{Unknown Questions Study}
%Then, we investigate what types of questions the LLMs do not know. 
We analyze the unknown and known questions of ChatGPT on question answering, arithmetic reasoning, and commonsense reasoning tasks across the six datasets. The known and unknown questions are determined based on the golden correctness label.

\paragraph{Knowledge Popularity}
We find that the LLM is prone to be ignorant of the atypical knowledge for openQA tasks. For example, when asked about the lyric writer of a less popular song, the model may produce different answers for differently verbalized questions as shown in Table~\ref{tab:intro_2_sample}. Besides, the most frequent answer is not always the correct one. 
We use the number of returned search results of Google and Bing as an indicator of the popularity of the knowledge for the question. In Table~\ref{tab:unknown_search_comparision}, we see the number of search results for unknown questions is significantly lower than for known questions. This suggests that the LLM has relatively poorer memorization of unpopular knowledge.

\begin{table}[t]
\centering
\small
\vspace{-0.0cm}
\begin{tabular}{@{}lcc@{}}
\toprule
Question Type & Google & Bing \\
\midrule 
Unknown & 7,497k & 1,255k  \\ 
Known & 10,929k & 2,647k \\ 
 \bottomrule
\end{tabular}
\vspace{-0.2cm}
\caption{The number of search results for unknown and known questions.}
\label{tab:unknown_search_comparision}
\vspace{-0.6cm}
\end{table}

\paragraph{Reasoning Steps}

\begin{table}[!th]
\small
\vspace{-0.2cm}
\begin{tabular}{p{7.2cm}}
\toprule
\texttt{Tom's restaurant gets 6 reservations a night. They normally order 2 meals that cost \$5 each and a \$5 bottle of wine. How much do they make a week if they are open 2 days a week?} \\
\midrule 
\texttt{A family wants to adopt for enviro-ethical reasons, what did they abhor?" 
(A) abandon; (B) foster child; (C) orphan; (D) biological child; (E) give away} \\
\bottomrule
\end{tabular}
\vspace{-0.2cm}
\caption{Two failed questions for ChatGPT that require longer reasoning steps.}
\label{tab:more_reasoning_example}
\vspace{-0.0cm}
\end{table}

For arithmetic reasoning and commonsense reasoning questions, if the solution requires more reasoning steps, or contains different arithmetic operations simultaneously, the model tends to confuse the order of operations. This leads to incorrect answers. As shown in the first example in Table~\ref{tab:more_reasoning_example}, the LLM needs to calculate the cost of a reservation first, which includes 2 meals with \$5 and a bottle of wine with \$5. Then calculate the cost of a night and a week. %However, the model calculates the number of meals and then calculates the price of the meals, which is not easy to calculate correctly. %This leads to an incorrect answer when calculating for the final step.

% \texttt{Tom's restaurant gets 6 reservations a night. They normally order 2 meals that cost \$5 each and a \$5 bottle of wine. How much do they make a week if they are open 2 days a week?}

% Besides, for arithmetic calculation, the model is more prone to make calculation errors on a 5-digit number or more, which reveals OpenAI does not request calculation API for the math problems.

For commonsense reasoning tasks, if the solution requires two or more reasoning steps, the model is more likely to make mistakes.
As shown in the second example in Table~\ref{tab:more_reasoning_example}, 
the model needs to reason the subject being concentrated on "adoption" first, and then "enviro-ethical reasons". 

\paragraph{Distracting Formulations}

\begin{table}[t]
\centering
\small
\vspace{-0.0cm}
\begin{tabular}{@{}lcc@{}}
\toprule
Question Type & Vicuna-13B & Llama2-13B \\
\midrule 
Unknown & 228.4 & 202.4 \\ 
Known & 204.0 & 185.1\\ 
 \bottomrule
\end{tabular}
\vspace{-0.2cm}
\caption{The negative log-likelihoods for unknown and known questions.}
\label{tab:unknown_neg_loglikelihood}
\vspace{-0.5cm}
\end{table}

When Distracting formulations exist, the model is prone to generate unexpected errors. We use "distracting" instead of "adversarial" to illustrate that the formulations are not crafted but are built-in.% which requires the model to focus on the chain of thought carefully and not to be distracted.

\begin{table}[!th]
\small
\vspace{-0.2cm}
\begin{tabular}{p{7.2cm}}
\toprule
\texttt{nell collects cards. she had 239 baseball cards and 38 10 cards. she gave some of her cards to jeff and now has 376 10 cards and 111 baseball cards left. how many more 10 cards than baseball cards does nell have?} \\
\midrule 
\texttt{The performer was ready to put on a show and stepped onto the launch platform, what was his job? (A) ocean; (B) battleship; (C) cape canaveral florida; (D) trapeze; (E) nasa} \\
\bottomrule
\end{tabular}
\vspace{-0.2cm}
\caption{Two questions with distracting formulations.}
\label{tab:distraction_example}
\vspace{-0.4cm}
\end{table}

As two examples shown in Table~\ref{tab:distraction_example}, the model should calculate the number of baseball cards that Neil has more than 10 cards, instead of being distracted by calculating how many cards Jeff has. 
The presence of "Cape Canaveral Florida" is a distractor compared to "trapeze" as the question mentions "launch platform".
% which easily leads the model to associate it with "rocket spacecraft astronaut". 

We report the negative log-likelihoods averaged across the six datasets of the known and unknown questions as the indicator of the atypical input in Table~\ref{tab:unknown_neg_loglikelihood}. We show that the unknown questions correlate with a higher score, i.e., higher atypicality.

\subsection{Impact of Diversified Questions}
\begin{figure}[t]
\vspace{-0.3cm}
    \begin{centering}
    \includegraphics[width=0.98\linewidth]{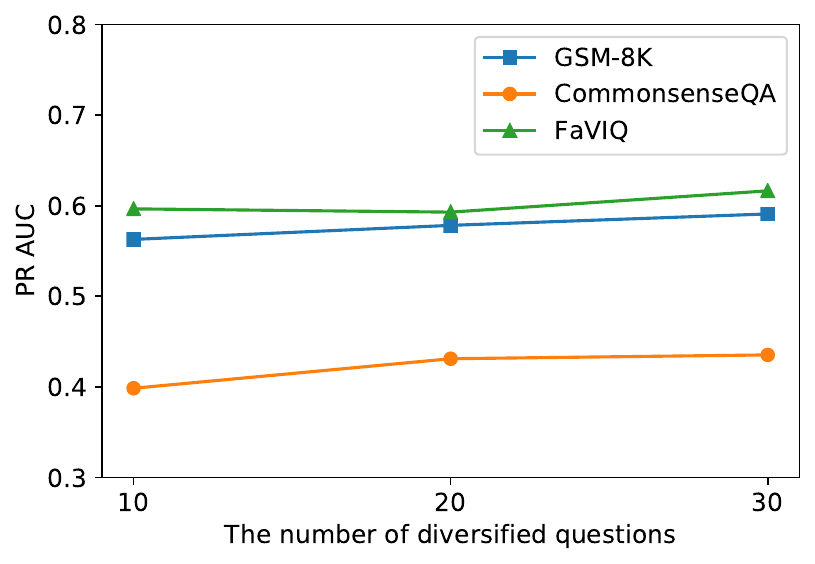}
    \end{centering}
    \vspace{-0.3cm}
    \caption{The performance of different numbers of diversified questions for the self-detection.}
    \label{fig:diff_q}
    \vspace{-0.3cm}
\end{figure}

We examine whether the number of paraphrased questions affects self-detection performance. Due to time and cost constraints, we only report the performance for ChatGPT on three representative datasets (FaVIQ, CommonsenseQA, and GSM8K) corresponding to the three tasks. We report the performance when the number of paraphrased questions is set to 10, 20, and 30. We observe that as the number of paraphrased questions increases, there is a slight improvement, as shown in Figure~\ref{fig:diff_q}. Our analysis reveals that some unknown questions may be answered coincidentally correctly when the number of questions is small. 
This inconsistency can be detected as the number of paraphrased questions increases. Additionally, for questions where the model is confident, the model continues to answer consistently, even with more questions. The two phenomena explain the improvement with more questions.

Finally, we conduct human evaluations on each sub-step of our self-detection in Appendix~\ref{subsec:component_eval} and report the costs when we call the OpenAI APIs in Appendix~\ref{subsec:costs}.

\section{Conclusion}
\label{sec:conclusion}
In this paper, we propose a simple yet effective method to self-detect whether an LLM generates non-factual responses, without referring to any other external resources. 
We conducted extensive experiments on four recent LLMs-- ChatGPT, GPT-4, Vicuna, and Llama 2 on three different types of tasks and demonstrated the effectiveness of our method and each two components. 
The two proposed components along with the existing methods can be combined for further utilization. 
Furthermore, we explore the question types that LLMs may struggle with, like low popularity and distracting formulations. 
Our method is applicable for continually upgrading LLMs. It can assist the LLMs to detect and improve their specific weaknesses, improving their reliability in the future.
%\newpage

\section*{Limitations}
While our method is effective, it still has several limitations.
% ambiguity and under-represented
Our self-detection method utilizes a model itself to diversify the verbalizations and thus the diversity is constrained by the LLM's abilities. In the future, we plan to collect more end-user questions from conversational agents or search engines to diversify the original questions to capture the built-in ambiguity of the questions. The ambiguity helps to further detect certain vulnerabilities of the model.
% Due to the time and cost constraints, we report the performance of 200 random samples from each test set for the six datasets. More data samples could be investigated for smoother experimental metrics. 
Besides, we detect the model's non-factuality through the divergence of the generated answers. It is unable to detect the cases where the model generates consistently but incorrectly, resulting the false negatives. Utilizing additional verifier LLMs or incorporating external knowledge for cross-checking is prevalent and we believe these would help to improve the detection performance. As this is not the focus of our paper, we omit the combinations with them.
% Besides, the diversity of the verbalized instructions may be limited compared with end-users as we collect the re-phrasings through LLMs or pre-defined rules. We plan to collect diversified questions from a wide range of end-users to construct test

\section*{Ethics Statement}
% We acknowledge the importance of the ACM code of Ethics and totally agree with it. 
We ensure that this work does not have explicit ethical considerations such as anonymity and privacy as all the models and datasets we use are public.
% We don't process the publicly available datasets for anonymity and privacy considerations and we are unclear about .
% We are unclear about anonymity and privacy in these datasets as we do not process the publicly available six datasets for these considerations.
We are unclear whether the publicly available LLMs may encode problematic bias as it is not the focus of this paper.
Our technique is used to detect what LLMs do not know and should not be used in other applications. 
At least for now, there is no risk of ethics for this method.

\section*{Acknowledgements}
 
This work was supported by 
the National Key R\&D Program of China with grant No.2020YFB1406704, 
the Natural Science Foundation of China (62272274, 61902219, 61972234),
the Natural Science Foundation of Shandong Province (ZR2021QF129). 

% We acknowledge that the publicly available datasets may contain information about per LLMs may encode problematic biases. It is unclear how the training process might interact with these problems.

% We ensure that this work is compatible with the Ethics code, in terms of the publicly accessed datasets and models.

% \section*{Acknowledgements}


\bibliography{custom}
\bibliographystyle{acl_natbib}

\appendix
\newpage
\appendix
\section{Appendix}
\label{sec:appendix}
\subsection{Prompts}
\label{sec:prompts}
We show the instruction to diversify the question verbalizations in Table~\ref{tab:paraphrase_prompt}. 
We recently optimized the instruction for the diversifying task by asking the model to provide 10 rephrased questions once. It helps to decrease the number of generations for the re-phrased questions to 1. The prompt is shown in Table~\ref{tab:new_rephrase_prompt} and we will report the performance with this new prompt in future work.

\begin{table}[!th]
\small
\vspace{-0.1cm}
\begin{tabular}{p{7.2cm}}
\toprule
\texttt{Given a question, paraphrase it to have different words and expressions but have the same meaning as the original question. Please note that you should not answer the question, but rather provide a re-phrased question.} \\
\bottomrule
\end{tabular}
\vspace{0.2cm}
\caption{The instruction for the diversifying task.}
\label{tab:paraphrase_prompt}
\vspace{-0.2cm}
\end{table}

%table 3
\begin{table}[!th]
\small
\vspace{-0.1cm}
\begin{tabular}{p{7.2cm}}
\toprule
\texttt{Paraphrase the input question to have different words and expressions but have the same meaning as the original question. Output the various paraphrases separated by '<br>'. Please note that you should not answer the question, but rather paraphrase it.} \\
\bottomrule
\vspace{-0.6cm}
\end{tabular}
\caption{The new instruction for diversifying.}
\label{tab:new_rephrase_prompt}
\vspace{-0.0cm}
\end{table}

The instruction for detecting wrong paraphrases of the question is in Table~\ref{tab:filter_prompt}.

% table 2
\begin{table}[!th]
\small
\vspace{-0.1cm}
\begin{tabular}{p{7.2cm}}
\toprule
\texttt{Determine whether the paraphrased question describes the same thing as the original question, and give "Contradicted" if they are not the same otherwise give "Same" as the result.} \\
\bottomrule
\vspace{-0.8cm}
\end{tabular}
\caption{The instruction for detecting wrong paraphrases.}
\label{tab:filter_prompt}
\vspace{-0.2cm}
\end{table}

\subsection{Evaluation of the Upgrading LLMs}
\label{subsec:evoluation}
We report the ratios of unknown questions for the continually upgrading models across the openQA, commonsense reasoning, and arithmetic reasoning tasks, where the unknown and known questions are determined by the golden correctness labels. As shown in Table~\ref{tab:unknown_ratios}. We see that GPT-4 performs the best and ChatGPT is weaker. Vicuna-13B and Lllam2-13B perform closely and both of them are weaker than the GPT series in terms of all tasks.

\begin{table}[t]
\small
\centering
\vspace{-0.3cm}
\begin{tabular}{@{}lcccc@{}}
\toprule
Dataset & ChatGPT & GPT4 & Vicuna & Llama 2  \\
\midrule 
ARC   & 0.10 & 0.05 & 0.57 & 0.36 \\ 
CSQA  & 0.19 & 0.13 & 0.47 & 0.34 \\ 
GSM8k & 0.11 & 0.05 & 0.64 & 0.65 \\ 
SVAMP & 0.15 & 0.07 & 0.44 & 0.43 \\ 
FaVIQ & 0.43 & 0.32 & 0.67 & 0.67\\ 
ComQA & 0.30 & 0.27 & 0.44 & 0.42 \\ 
 \bottomrule
\end{tabular}
\caption{Comparison of the ratios of unknown questions for different LLMs. CSQA is commonsenseQA for short.}
\vspace{-0.2cm}
\label{tab:unknown_ratios}
\vspace{-0.0cm}
\end{table}

\subsection{Additional Experiments}
\label{subsec:additional_exp}
\paragraph{Ablation on Diversifying Verbalizations}
We also report the performance when we combine ConsistAns (divergence of randomly sampled answers without using diversifying verbalizations) with Atypicality in Table~\ref{table:consistency_performance}. It falls behind our SelfDetection in most datasets, which again reveals the necessity of incorporating verbalizations for the detection task.

\begin{table*}[!t]
\centering\small
\begin{tabular}{lp{3.5em}<{\centering}p{6.5em}<{\centering}p{6em}<{\centering}p{3.5em}<{\centering}p{3.5em}<{\centering}p{3.5em}<{\centering}p{3.5em}<{\centering}}
\toprule
& ARC & CommonsenseQA & GSM-8K &	SVAMP & FaVIQ & ComQA  \\
\midrule
\multicolumn{4}{@{}l}{\emph{Vicuna-13B}}\\
SelfDetection   & {54.55} & {62.93} & {53.31} &  {71.19}   & {39.45} & {66.97} \\
ConsistAns + Atypicality	    & 51.32 & 58.66 & 53.12 & 67.43 & 37.33 & 60.04 \\
\midrule

\multicolumn{4}{@{}l}{\emph{Llama2-13B}}\\
SelfDetection  & {77.73} & {71.95} & {50.38} &  {70.33}   & 39.83 & 52.36 \\
ConsistAns + Atypicality    & 78.22 & 68.61 & 51.92 & 68.90 & 40.24 & 54.11 \\
\bottomrule

\end{tabular}
\caption{The PR-AUC of SelfDetection (Consistency + Atypicality) and ConsistAns + Atypicality on Vicuna-13B and Llama2-13B.}
\label{table:consistency_performance}
\vspace{-0.2cm}
\end{table*}

\paragraph{Linear Combination of Our Components} In this paper, our method contains two components (using consistency, entropy, atypicality, and its normalized version specifically). We conduct a simple linear combination where each feature is normalized to [0, 1] and the weights are all set to 1. The performance is shown in Table~\ref{table:other_combinatioin}. We see the performance is close to the fitted version using XGBoost.

\begin{table*}[!t]
\centering\small
\begin{tabular}{lp{3.5em}<{\centering}p{6.5em}<{\centering}p{6em}<{\centering}p{3.5em}<{\centering}p{3.5em}<{\centering}p{3.5em}<{\centering}p{3.5em}<{\centering}}
\toprule
& ARC & CommonsenseQA & GSM-8K &	SVAMP & FaVIQ & ComQA  \\
\midrule
\multicolumn{4}{@{}l}{\emph{Vicuna-13B}}\\
SelfDetection   & {54.55} & {62.93} & {53.31} &  {71.19}   & {39.45} & {66.97} \\
SelfDetection (LC)	& 53.16 & 59.37	 & 51.34 & 68.15 & 37.65 & 61.60 \\
\midrule

\multicolumn{4}{@{}l}{\emph{Llama2-13B}}\\
SelfDetection  & {77.73} & {71.95} & {50.38} &  {70.33}   & 39.83 & 52.36 \\
SelfDetection (LC)    & 72.51 & 70.21 & 49.38 & 68.39 & 37.84 & 52.02 \\
\bottomrule

\end{tabular}
\caption{The PR-AUC of SelfDetection (fitting using XGBoost on dev set) and SelfDetection (LC) (LC is the linear combination for short).}
\label{table:other_combinatioin}
\vspace{-0.2cm}
\end{table*}

\subsection{Component Evaluation} 
\label{subsec:component_eval}
We analyze the precision of each component in our framework. For the first paraphrase module, we randomly sampled 100 paraphrases generated from the four LLMs. Then we manually label whether the rephrased versions describe the same thing as the original questions. We report the human-labeled agreement ratio upon the 100 instances as the paraphrase precision. 

The precision for the commonsense reasoning tasks is 100\% as we only exchange the options as the paraphrased version. In arithmetic reasoning tasks, the precision is 99\% as we only exchange the subjects of the question for a paraphrased version, with the remaining 1\% errors due to the conflicts between animal names and human names. 
For openQA questions, the precisions for ChatGPT, GPT-4, Vicuna-13B, and Llama2-13B are 95\%, 95\%, 93\%, and 93\% respectively. 

Then, we evaluate the answer clustering performance directly and omit evaluating the consistency detection performance, as we group the answers solely based on whether the two answers are consistent. The precision is measured by calculating the proportion of answer-pairs in the intersection correctly assigned between the output cluster $\Omega = \{ \omega_1,  \ldots, \omega_k \}$ and the ground-truth cluster $\mathcal{C} = \{ {c_1}, \ldots, {c_p} \}$. We report the clustering precision in our manually labeled 400 clusters. 
\begin{equation*}
\text{Precision}(\mathcal{C}, \Omega)= \frac{1}{k}\sum_{i=1}^k \frac{ \binom{\max \limits_{j} |\omega_j \cap c_i|}{2} } { \binom{|c_i|}{2} },
\end{equation*}

We achieved 100\% precision for the commonsense reasoning task for the four LLMs. For openQA questions, we achieve precisions of 89\%, 90\%, 83\%, and 81\% for ChatGPT, GPT-4, Vicuna-13B and Llama2-13B respectively. For arithmetic reasoning tasks, the precision scores are 92\%, 93\%, 89\%, and 88\% for ChatGPT, GPT-4, Vicuna-13B and Llama2-13B respectively.

\subsection{Costs}
\label{subsec:costs}

\begin{table}[t]
\centering
\small
\vspace{-0.0cm}
\begin{tabular}{@{}lccc@{}}
\toprule
Methods & QA & CSQA & Arith.  \\
\midrule 
\textbf{ChatGPT (gpt-3.5-turbo)} &  &  & \\
TP \& PRL & 0.00008 & 0.0002 &  0.00006 \\ 
SCGPT \& CA & 0.002 & 0.004  & 0.0006 \\ 
SelfDetect & 0.004 & 0.004  & 0.0006 \\ 
% \midrule
% \textbf{text-davinci-003} &  &  & \\
% TP \& PRL & 0.0008 & 0.0014 &  0.0008 \\ 
% SCGPT \& CA & 0.022 & 0.038  & 0.008 \\ 
% SelfDetect & 0.049 & 0.038  & 0.008 \\ 
\midrule
\textbf{GPT-4} &  &  & \\
TP \& PRL & 0.0024& 0.0068 &  0.0014 \\ 
SCGPT \& CA &  0.046  & 0.105   & 0.014 \\ 
SelfDetect & 0.092 & 0.106  & 0.014 \\ 
 \bottomrule
\end{tabular}
\caption{The costs per question for the TokenProbs (TP), Perplexity(PRL), ConsistAnswers (CA), SelfCheckGPT (SCGPT) and SelfDetection methods on OpenQA (QA), CommonsenseQA (CSQA) and arithmetical reasoning (Arith.) tasks.}
\label{tab:costs}
\vspace{-0.2cm}
\end{table}

We report the costs for our self-detection and the compared methods.
For open-source models like Vicuna, we deploy them ourselves for inference. For those close-sourced like ChatGPT, we request APIs. The costs per question in U.S. dollars across different tasks are shown in Table~\ref{tab:costs}.
% \section{Example Appendix}
% \label{sec:appendix}

% This is a section in the appendix.

\end{document}